%% file: root.tex
\documentclass[letterpaper, 10 pt, conference]{ieeeconf}  

\IEEEoverridecommandlockouts                              

\makeatletter
\let\NAT@parse\undefined
\makeatother
\usepackage[numbers,sort&compress]{natbib}

\usepackage[pdftex]{graphicx}
\usepackage{amsmath}
\usepackage{amssymb}  
\usepackage{subfigure}
\usepackage{subcaption}
\usepackage{multirow}
\usepackage{array,booktabs}
\usepackage{diagbox}
\usepackage{balance}
\usepackage{xspace}
\usepackage{algorithm}
\usepackage{algpseudocode}
\usepackage{adjustbox}
\usepackage{romannum}

\usepackage[final]{hyperref}
\hypersetup{
 colorlinks=true,
 linkcolor=blue,
 filecolor=magenta,
 urlcolor=blue,
 citecolor=black
}
\usepackage{cleveref}
\crefname{figure}{Fig.}{Figs.}
\Crefname{figure}{Fig.}{Figs.}
\usepackage[font=small]{caption}
\usepackage{rpm_SIunits}
\usepackage{rpm_acronyms}
\usepackage{rpm_math}
\usepackage{rpm_misc}
\usepackage{bbm}

\usepackage{textcomp}

\usepackage{hyperref}
\hypersetup{
    colorlinks=true,
    linkcolor=blue,
    filecolor=magenta,      
    urlcolor=cyan,
}

\usepackage{soul,color}
\usepackage{lipsum}
\usepackage{dsfont}

\usepackage{xcolor}
\usepackage{colortbl}

\usepackage{xparse}

\usepackage{amssymb}
\usepackage{pifont}
\usepackage{tikz} 
\usepackage{subcaption}
\usepackage{caption}
\title{\LARGE \bf Clutt3R-Seg: Sparse-view 3D Instance Segmentation \\ for Language-grounded Grasping in Cluttered Scenes 
}     

\author{
\makebox[\textwidth][c]{%
    Jeongho Noh${}^{1}$, 
    Tai Hyoung Rhee${}^{1}$, 
    Eunho Lee${}^{2}$, 
    Jeongyun Kim${}^{1}$, 
    Sunwoo Lee${}^{3}$ and 
    Ayoung Kim${}^{1*}$%
}%
\thanks{$^\dagger$This work was supported by Hyundai Motor Company and Kia, and by the Institute of Information \& communications Technology Planning \& Evaluation (IITP) grant funded by the Korea government(MSIT) No.2022-0-00480, Development of Training and Inference Methods for Goal-Oriented Artificial Intelligence Agents.}
\thanks{$^{1}$J. Noh, TH. Rhee, J. Kim and A. Kim are with the Dept. of Mechanical Engineering, SNU, Seoul, S. Korea {\tt\small [shwjdgh3842, williamrhee, jeongyunkim, ayoungk]@snu.ac.kr}}%
\thanks{$^{2}$E. Lee is with the Interdisciplinary Program in Artificial Intelligence, SNU, Seoul, S. Korea {\tt\small \{eunho1124\}@snu.ac.kr}}%
\thanks{$^{3}$S. Lee is with the Robotics Lab, Hyundai Motor Company, S. Korea {\tt\small \{twosun\}@hyundai.com}}%
}

\begin{document}

\maketitle
\thispagestyle{empty}
\pagestyle{empty}

\input{0_abstract}
\input{1_introduction}
\input{2_relatedwork}

\input{3_method}
\input{4_experiment}
\input{5_conclusion}


\small
\bibliographystyle{IEEEtranN} 
\bibliography{rpm_string,references}

\end{document}

%% file: 0_abstract.tex
\begin{abstract}


Reliable 3D instance segmentation is fundamental to language-grounded robotic manipulation. Its critical application lies in cluttered environments, where occlusions, limited viewpoints, and noisy masks degrade perception. 
To address these challenges, we present Clutt3R-Seg, a zero-shot pipeline for robust 3D instance segmentation for language-grounded grasping in cluttered scenes.
Our key idea is to introduce a hierarchical instance tree of semantic cues. Unlike prior approaches that attempt to refine noisy masks, our method leverages them as informative cues: through cross-view grouping and conditional substitution, the tree suppresses over- and under-segmentation, yielding view-consistent masks and robust 3D instances.
Each instance is enriched with open-vocabulary semantic embeddings, enabling accurate target selection from natural language instructions.
To handle scene changes during multi-stage tasks, we further introduce a consistency-aware update that preserves instance correspondences from only a single post-interaction image, allowing efficient adaptation without rescanning. Clutt3R-Seg is evaluated on both synthetic and real-world datasets, and validated on a real robot. Across all settings, it consistently outperforms state-of-the-art baselines in cluttered and sparse-view scenarios. 
Even on the most challenging heavy-clutter sequences, Clutt3R-Seg achieves an AP@25 of 61.66, over 2.2× higher than baselines, and with only four input views it surpasses MaskClustering with eight views by more than 2×.
The code is available at: \href{https://github.com/jeonghonoh/clutt3r-seg}{https://github.com/jeonghonoh/clutt3r\-seg}.

\end{abstract}

%% file: 1_introduction.tex
\section{Introduction}
\label{sec:intro}

Robotic manipulation is widely adopted in automation and industrial applications. Still, there exists a major challenge in severely cluttered scenes where heavy occlusions degrade instance segmentation and reliable perception. Early works~\cite{ornek2023supergb, xie2022rice, xie2021unseen} exploring a single RGB-D input were often vulnerable under limited viewpoints or heavy occlusion. As a mitigation, recent methods~\cite{yin2024sai3d, ye2024gaussian, zhao2025sam2object} adopt dense inputs, but fail to update after scene changes such as object removal or displacement, even with the increased computational cost.
In this regard, securing accurate and efficient 3D segmentation from sparse views remains a significant challenge.


Another major challenge with cluttered scenes is that semantic features are often ambiguous, making language-grounded target identification difficult. Recent works~\cite{wu2024opengaussian, shi2024language} addressing this challenge embed per-view features into a 3D scene representation to encourage cross-view consistency. However, because these pipelines still depend on per-frame masks, over- and under-segmentation are inherited by the 3D representation and break cross-view consistency. Another approach,  LangSplat~\cite{qin2024langsplat}, employs hierarchical features to enrich semantics. Nevertheless, maintaining per-object, per-part feature maps introduces high computational overhead, while semantics still remain unstable across views under heavy clutter, leading to failures in language-grounded grasping.

\begin{figure}[!t]
    \centering
    \includegraphics[width=0.98\linewidth]{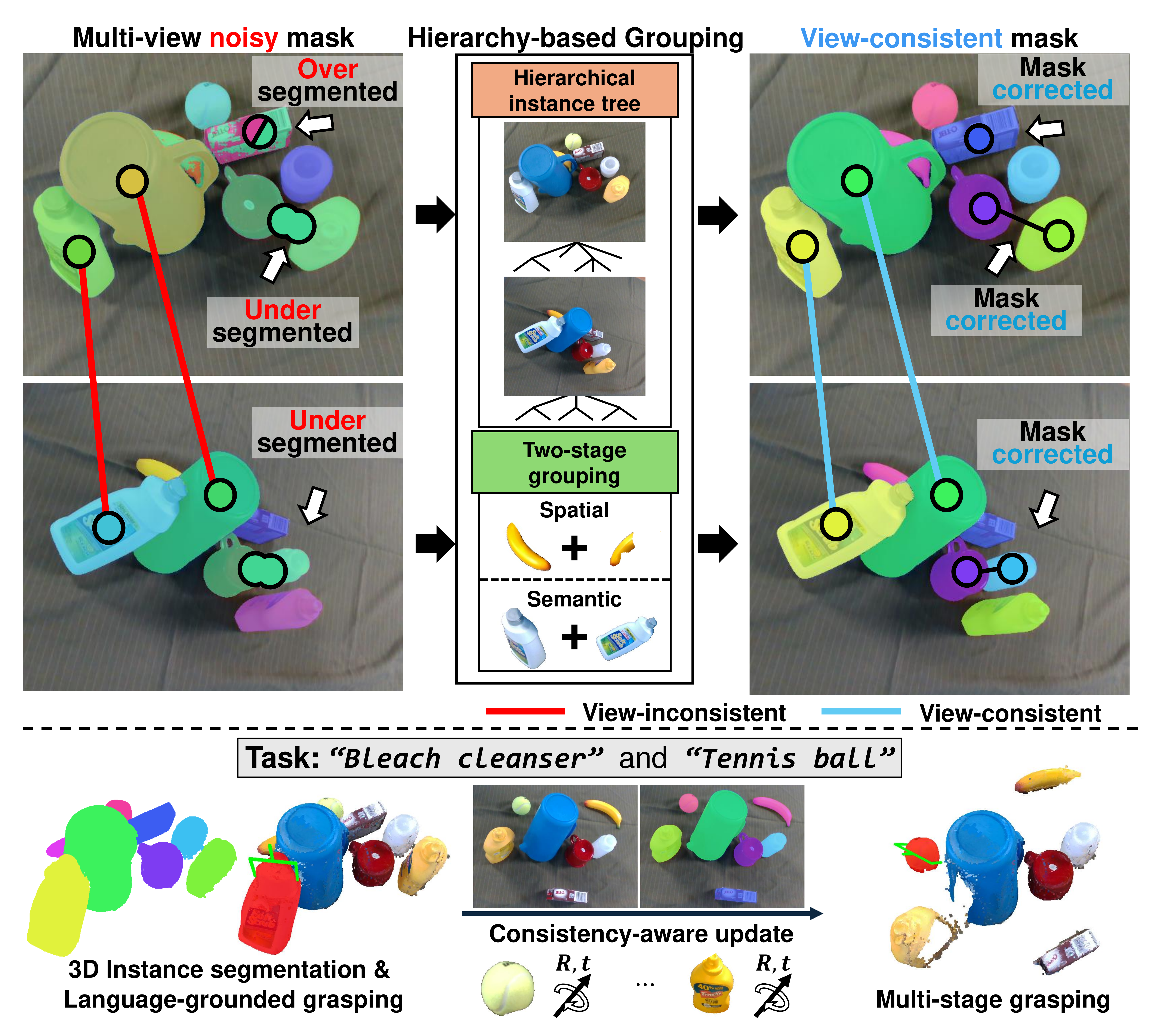}
    \caption{\textbf{Clutt3R-Seg} resolves over- and under-segmentation and cross-view inconsistencies from erroneous noisy masks in sparse-view cluttered scenes through a hierarchy-based instance mask grouping algorithm, yielding view-consistent masks and robust 3D instance segmentation. These consistent 3D instances enable language-grounded target identification and consistency-aware multi-stage grasping by detecting displaced objects and optimizing their poses after interaction.}
    \label{fig:1}
    \vspace{-6mm}
\end{figure}

\begin{figure*}[!t]
    \centering
    \includegraphics[width=0.90\linewidth]{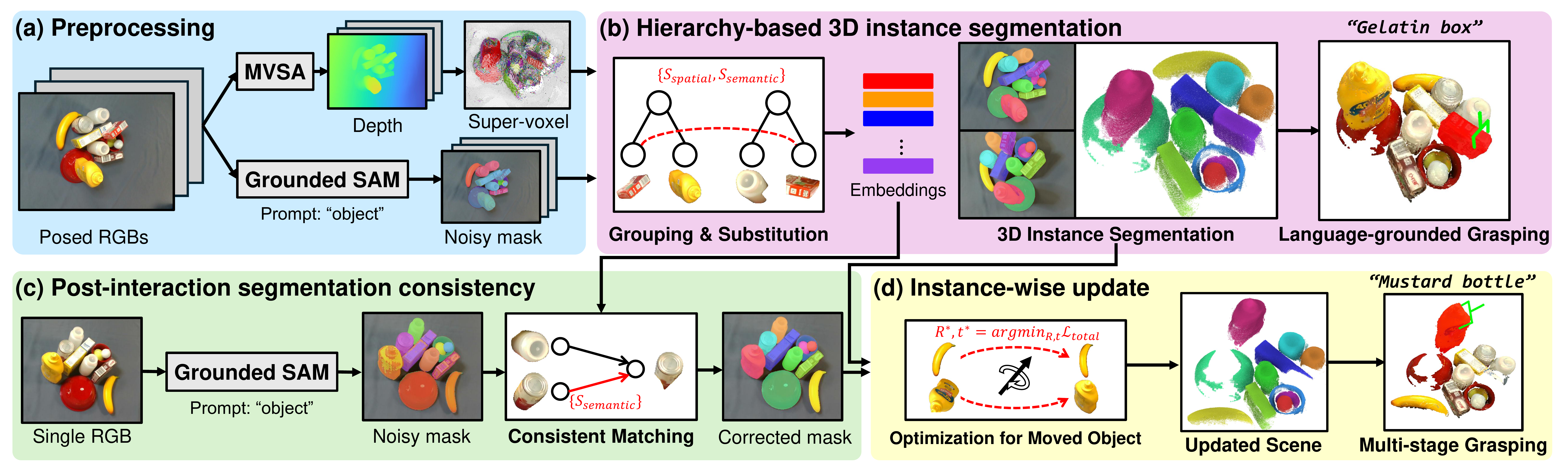}
    \caption{Overview of Clutt3R-Seg pipeline. \textbf{(a)} From posed sparse-view RGB inputs, we estimate depth \cite{izquierdo2025mvsanywhere} and obtain noisy instance masks \cite{ren2024grounded} with the prompt ``object". \textbf{(b)} A hierarchy-based grouping yields view-consistent instance mask groups robust to over-/under-segmentation, enabling reliable 3D instance segmentation in cluttered scenes. Enriched semantic embedding allows text-based target identification. \textbf{(c)} A single post-interaction image is associated with prior instances to preserve segmentation consistency. \textbf{(d)} The system detects the new target and displaced objects, optimizing their rigid transformation via a differentiable loss for reliable multi-stage grasping.}
    \label{fig:2}
    \vspace{-7mm}
\end{figure*}

In this paper, we present Clutt3R-Seg, a pipeline for zero-shot 3D instance segmentation and consistency-aware multi-stage grasping that enables language-grounded manipulation in heavily cluttered, sparse-view settings. 
As shown in Fig.~\ref{fig:1}, our key idea is to organize noisy 2D mask candidates into a hierarchical structure that reveals redundancy among overlapping masks. 
Through cross-view grouping of high-similarity leaves and conditional substitution, our method mitigates over- and under-segmentation and yields view-consistent 2D masks. These masks in turn support robust 3D instance segmentation under heavy clutter.
Each instance is further enriched with open-vocabulary semantic embeddings, enabling accurate language-grounded target selection and grasping. Moreover, by preserving instance consistency from a single post-interaction image, the system remains reliable under significant scene changes, supporting efficient multi-grasp execution. Our main contributions are as follows:

\noindent
\begin{itemize}
\item \textbf{A Novel Hierarchy-based 3D Instance Segmentation:} By leveraging a hierarchical instance tree, our method organizes noisy mask candidates into reliable cross-view instance groups, enabling robust 3D instance segmentation even under sparse multi-view inputs.

\noindent
\item \textbf{Language-grounded Target Identification:} The semantically enriched features obtained from reliable sparse multi-view correspondences of our consistent 3D instances enable direct grounding of high-level language commands, allowing the robot to identify and localize a specific target object from a text prompt for grasping.

\noindent
\item \textbf{Efficient Update Method for Multi-stage Grasping:} After each grasp, the proposed update method relies solely on a single post-interaction image to re-identify the remaining instances and recognize displaced objects to maintain consistency in the scene representation.

\noindent
\item \textbf{Extensive Robotic Validation and Public Release:} We validate the entire system through extensive experiments in both the synthetic and the real-world settings, demonstrating its robustness and generalization. To ensure reproducibility, we will publicly release all our source code, simulation environment, and synthetic datasets.

\end{itemize}

%% file: 2_relatedwork.tex
\section{Related Work}
\label{sec:relatedwork}

\subsection{Multi-view 3D Instance Segmentation}


Multi-view 3D instance segmentation often requires consistent object representations across viewpoints, which is generally challenging to obtain. To overcome this requirement, affinity-based methods~\cite{yin2024sai3d} build similarity matrices over 2D masks via region growing, while others incorporate multi-view features or geometric cues for unified 3D segmentation~\cite{nguyen2023open3dis, yang2023sam3d, xu2025sampro3d}. More recently, MaskClustering~\cite{yan2024maskclusteringviewconsensusbased} constructs a global mask graph whose edges are weighted by a view-consensus rate and merges masks through iterative clustering. However, these approaches inherently assume high-quality per-view masks. In cluttered and occluded scenes, 2D segmentation is noisy, prone to over- and under-segmentation, so the affinity matrix becomes unreliable~\cite{yin2024sai3d, yan2024maskclusteringviewconsensusbased}. GraphSeg~\cite{tang2025graphsegsegmented3drepresentations} targets sparse inputs by constructing dual correspondence graphs from 2D feature matches and inferred 3D structure, then merging fragmented masks via edge addition and graph contraction. While effective, its dependence on initial 2D mask quality leaves it sensitive to under- and over-segmentation, which destabilize the graph and degrade later stages. 
In contrast, we introduce a hierarchy-based mask grouping algorithm that constructs per-view instance trees and performs bottom-up grouping, stabilizing instance segmentation across views and yielding consistent 3D instances under sparse observations and heavy clutter.

\subsection{Language-grounded Target Identification and Grasping}
Recent advancements in robotic manipulation have increasingly leveraged vision–language models for language-grounded interaction in complex environments. ConceptGraphs~\cite{gu2024conceptgraphs} converts posed RGB-D images into an open-vocabulary, object-centric 3D scene graph by segmenting images, fusing multi-view regions into object nodes, and using vision-language models to name objects and infer relations. Alternatively, GaussianGrasper~\cite{zheng2024gaussiangrasper3dlanguagegaussian} embeds language features directly into 3D Gaussian primitives to enable language-guided grasping. Similarly, GraspSplats~\cite{ji2024graspsplatsefficientmanipulation3d} constructs feature-enhanced 3D Gaussians with hierarchical descriptors, facilitating efficient part-level grasp queries and multi-stage grasping via point tracking. However, because these methods only depend on noisy 2D masks, their performance can degrade under heavy clutter. In addition, scene update module requires densely consecutive frames, limiting use with sparse or non-sequential captures. 
In contrast, Clutt3R-Seg resolves under- and over-segmentation by organizing initial noisy masks with a hierarchical instance tree, recovering accurate masks, and enriching semantics for robust target grasping in clutter. It also supports multi-stage grasping under sparse frames and significant scene changes, by accurate instance correspondences via robust and consistent semantic features.

%% file: 3_method.tex
\section{Methodology}

Clutt3R-Seg comprises three stages: preprocessing \secref{method:pre}, \textit{\textbf{hierarchy-based 3D instance segmentation}} \secref{sec:inst-seg}, and \textit{\textbf{consistency-aware scene update}} \secref{sec:multi-grasp}, leading to successful \textit{\textbf{multi-stage grasping}} as illustrated in \figref{fig:2}.

\begin{figure}[!t]
    \centering
    \includegraphics[width=0.95\linewidth]{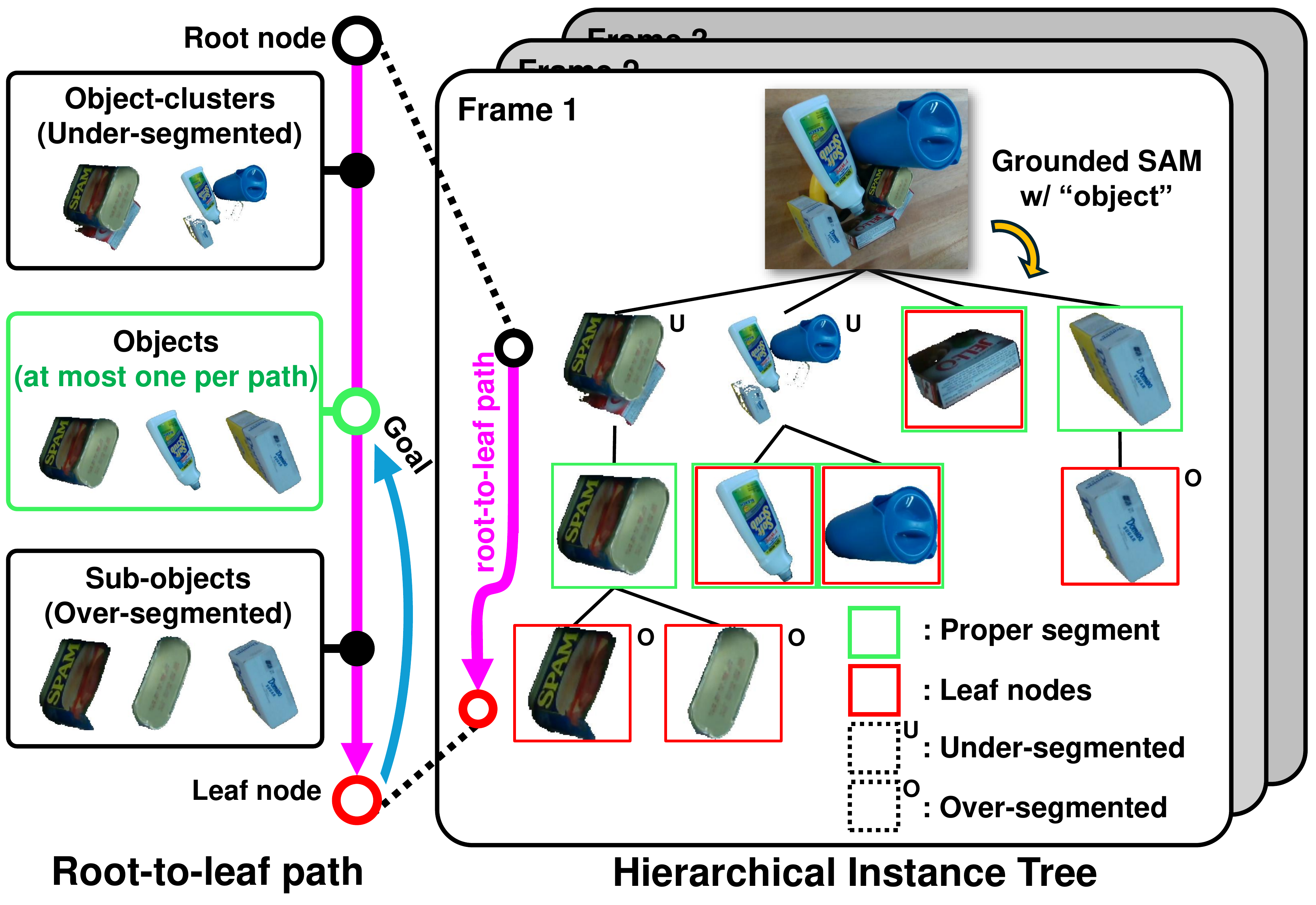}
    \caption{\textbf{Hierarchical structure of masks.} Grounded SAM outputs masks for all object-clusters, objects and sub-objects corresponding to under-segmentation, proper segmentation and over-segmentation, respectively. Such noisy masks in cluttered scenes are organized into hierarchical trees, assigning a \textit{child node} to a \textit{parent node} if its mask is contained within the pixel level. In this hierarchy, \textbf{at most one proper segment exists per root-to-leaf path}, enabling bottom-up grouping of leaf nodes across frames for instance consistency.}
    \label{fig:3}
    \vspace{-6mm}
\end{figure}

\subsection{Preprocessing}
\label{method:pre}
Given posed RGB images, we first generate initial 2D masks by prompting Grounded SAM~\cite{ren2024grounded} with the single token ``object". We adopt this generic prompt because the set of present categories is unknown, which removes per-scene prompt engineering while covering arbitrary, unseen objects. These masks contain all proper, under- and over-segmented masks. In parallel, we reconstruct a 3D point cloud from the same images using a depth estimation model (MVSAnywhere~\cite{izquierdo2025mvsanywhere} in our implementation). To control computation and memory, we apply voxel grid downsampling, retaining per-voxel averaged appearance and geometry.

After obtaining the downsampled point cloud and masks, we cluster geometrically homogeneous regions into compact spatial primitives as in \figref{fig:2}(a), termed super-voxels. Super-voxels are formed via k-NN clustering based on a combination of Euclidean distance and surface normal cosine similarity between two adjacent points $i$ and $j$, defined as:

\begin{equation}
\label{eq:supervoxel}
d_{ij}=\alpha \left\| x_i-x_j\right\|_2+\beta(1-\left< n_i,n_j \right>)
\end{equation}

\noindent with $x_i,x_j$ and $n_i,n_j$ representing the 3D coordinate and normal vector of points $i$ and $j$, respectively. This formulation encourages grouping of neighboring points with consistent surface orientation, leading to geometrically similar regions that enable robust and efficient cross-view grouping and 3D instance segmentation.

\subsection{Hierarchy-based 3D Instance Segmentation}
\label{sec:inst-seg}
\begin{figure}[!t]
    \centering
    \includegraphics[width=0.85\linewidth]{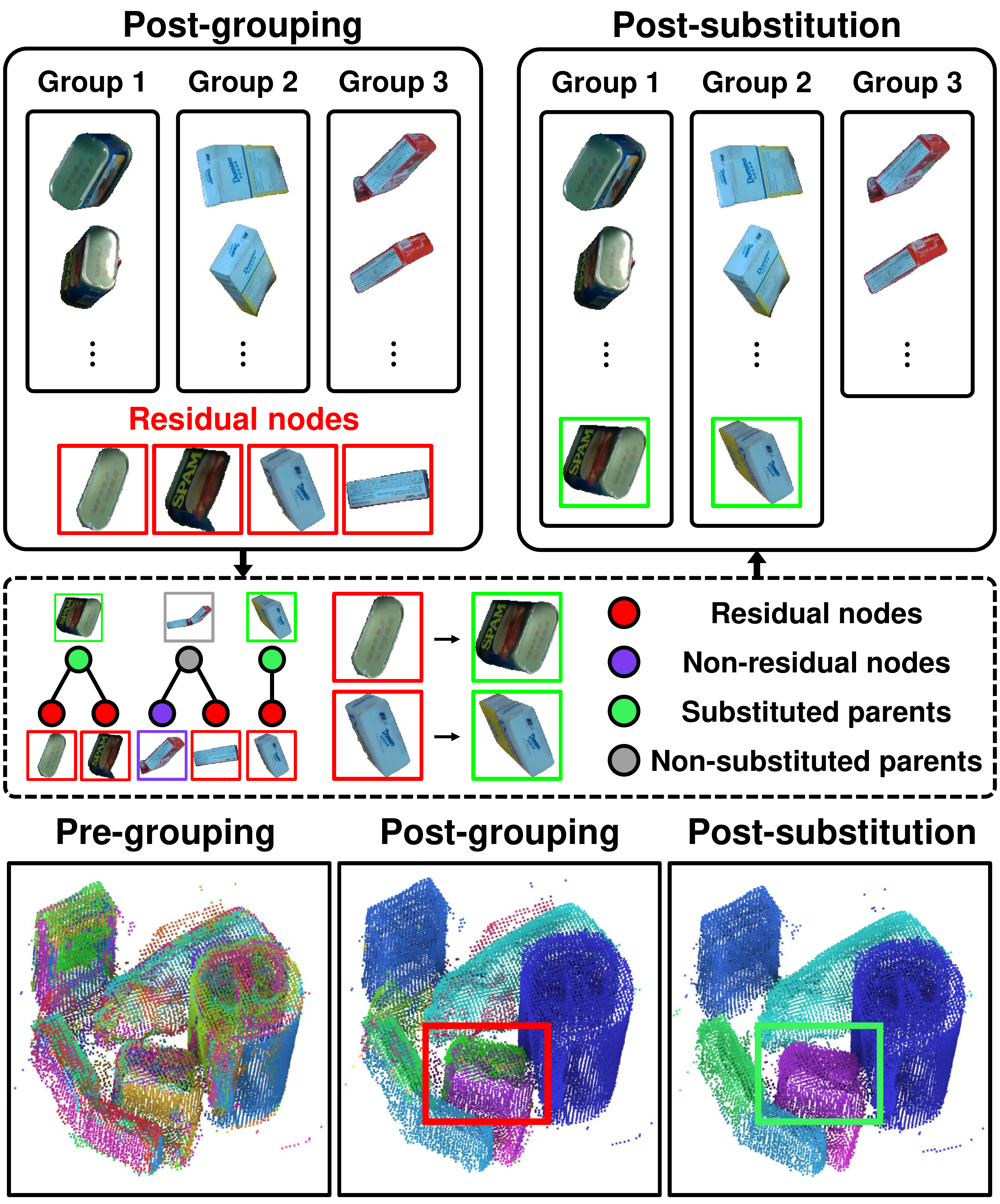}
    \caption{\textbf{Outline of mask grouping and substitution.} Initially inconsistent instance masks are properly organized to their corresponding instance groups via hierarchy-based instance mask grouping (\algoref{algo:1}), and further refined by re-grouping via residual-node parent substitution, resulting in properly segmented instances. The \textcolor{red}{red box} illustrates an over-segmentation before substitution, while the \textcolor{green}{green box} shows the corrected segmentation after substituting the residual node with its parent.}
    \label{fig:4}
    \vspace{-7mm}
\end{figure}

Our zero-shot hierarchy-based 3D instance segmentation module aims to achieve robust, view-consistent instance segmentation in sparse-view, cluttered scenes with severe occlusions.
The over- and under-segmentation errors from Grounded SAM are stochastically irregular and do not consistently appear across multiple views. Motivated by this empirical observation, we construct a hierarchical instance tree as in \figref{fig:3}, organizing the initial masks containing proper, over- and under-segmentation. Subsequently, hierarchy-based instance mask grouping (\algoref{algo:1}) groups masks that belong to the same instance. Residual-node parent substitution then corrects the remaining residual nodes, as illustrated in \figref{fig:4}, leading to accurate 3D instance segmentation.

\input{alg/algorithm1}

\noindent \textbf{Notations.} 
A graph $G$ consists of leaf nodes $V$, with each edge $E_{u,v}$ separately encoding spatial similarity $S_{\text{spatial}}$ and semantic similarity $S_{\text{semantic}}$, each with a threshold $\tau_{spat}$ and $\tau_{sem}$, respectively, between an unordered pair of leaf nodes $(u,v)$. The function $\phi(\cdot)$ maps nodes to the corresponding frame index to prevent the generation of edges sharing the same root. Here, $\mathcal{N}(u)$ denotes the set of neighbors directly connected to node $u$, while \textsc{Set}$(u)$ represents the collection of original nodes that compose the grouped node $u$.

\noindent
1) \textbf{Instance Tree Construction. }
To refine initial instance masks, we construct a hierarchical instance tree by defining parent–child relationships based on object-level mask hierarchy (object-cluster, object, sub-object), as shown in \figref{fig:3}.
Key intuition is that along each root-to-leaf path, at most one mask corresponds to the proper segment. If a correct object-level mask exists, any larger mask that contains it reflects under-segmentation, while any smaller mask nested inside it reflects over-segmentation. By enforcing this \emph{one-proper-per-path constraint}, the tree provides a principled way to distinguish correct instances from erroneous masks, yielding a robust structure for subsequent cross-view mask grouping.

Since each root-to-leaf path contains at most one proper segment, it can be regarded as the proper segment candidate. Because each path is uniquely determined by its leaf node, the candidate set reduces to the leaf nodes. Therefore, we treat the leaf nodes as proper segment candidates and use them as the basis for cross-view instance grouping.

\noindent
2) \textbf{Cross-view Instance Grouping. }
With the instance trees constructed for each view, we aggregate leaf nodes across views and group them into object-level instances. To remain robust under clutter and sparse-view conditions, we adopt a two-stage grouping strategy, as detailed in \algoref{algo:1}. In particular, we use \emph{spatial similarity} both to resolve ambiguities among same category duplicates and to stabilize grouping under sparse multi-view inputs with appearance shifts.
Within $E_{u,v}$, we define $S_{\text{spatial}}$ as a weighted Jaccard index based on super-voxel occupancy, quantifying the spatial overlap in 3D space. Let $\mathcal{K}$ be the set of all super-voxels in the scene, $o_i(k)\in[0,1]$ be the occupancy ratio of super-voxel $k\in\mathcal{K}$ by mask $i$. Each super-voxel has a normalized weight $\hat{w}_k$ proportional to the number of points it contains. The spatial similarity is:
\begin{equation}
\label{eq:S_spatial}
S_{\text{spatial}}(i, j) = \frac{\sum_{k \in \mathcal{K}} \hat{w}_k \min(o_i(k), o_j(k))}{\sum_{k \in \mathcal{K}} \hat{w}_k \max(o_i(k), o_j(k)) + \epsilon}.
\end{equation}
The denominator and numerator denote weighted union and intersection, with a small constant $\epsilon$ for numerical stability.

$S_\text{semantic}$ is defined as the cosine similarity between embeddings from Duoduo CLIP~\cite{lee2025duoduo}, offering a viewpoint-robust text–image space effective for cross-view mask grouping:

\begin{equation}
\label{eq:S_sem}
S_{\text{semantic}}(i, j) = \langle \mathbf{e}_i, \mathbf{e}_j \rangle
\end{equation}
\noindent where $\mathbf{e}_k$ represents embeddings of input mask $k$. 

Duoduo CLIP allows multi-view input, which supports robust cross-view correspondences under various viewpoints and appearance variations.
It further enables a mask group of the same instance to aggregate evidence across views to provide a more robust representation of the instance, facilitating language-grounded target identification and maintaining segmentation consistency under severe drift.

Similarities are evaluated for all candidate node pairs, and grouping is performed iteratively by selecting the pair with the highest similarity score, first under $S_{\text{spatial}}$ until no edges exceed the spatial threshold $\tau_{spat}$, and then under $S_\text{semantic}$ to group the remaining nodes. As shown in \figref{fig:4}, two or more leaf nodes that satisfy the similarity criterion are grouped.

\noindent
3) \textbf{Residual-node Parent Substitution.}
Leaf nodes that remain ungrouped are referred to as residual nodes $S$. They often arise from over-segmentation, where the residual node masks cover only a small part of an object, resulting in low $S_{\text{spatial}}$ and $S_{\text{semantic}}$ due to limited geometric coverage and contextual evidence, which leads to grouping failure. This over-segmentation issue, compounded by the aforementioned under-segmentation, presents a key challenge of discerning whether a residual should be attributed to over- or under-segmentation. To address this, we introduce a \emph{conditional leaf-to-parent substitution and re-grouping procedure}. A residual node is substituted with its parent node only when all of the parent's other descendants are residual nodes, thereby enforcing the constraint illustrated in Fig.~\ref{fig:3} that at most one proper segment exists along each root-to-leaf path. Empirically, the process converges within a single iteration, resulting in leftover residual nodes properly grouped into their corresponding instances.

\noindent
4) \textbf{Super-voxel Majority Vote. }
With mask grouping complete, we project the 2D masks to 3D super-voxels to perform 3D instance segmentation. Each converged node group defines a view-consistent 3D instance, and we define its confidence with the group size, reflecting the reliability of multiple constituent masks. We regard each group as a separate instance, which is then allocated to super-voxels through a voting scheme weighted by the point coverage of the grouped masks. This majority voting ensures that geometrically confident super-voxels have greater influence, yielding a more reliable 3D instance segmentation.

\subsection{Language-grounded Multi-stage Grasping}
\label{sec:multi-grasp}
\begin{figure}[!t]
    \centering
    \includegraphics[width=0.8\linewidth]{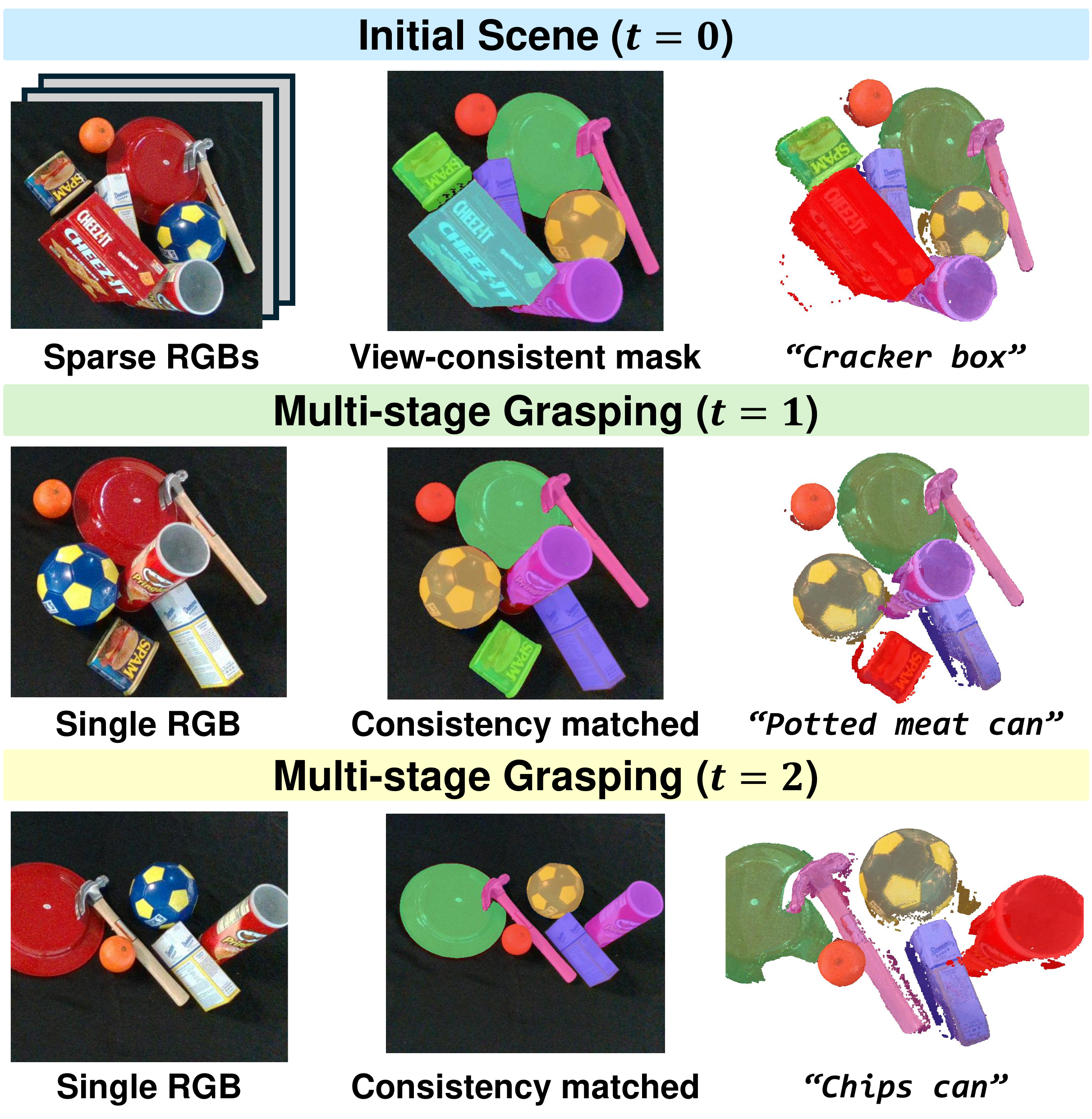}
    \caption{\textbf{Result of multi-stage target identification and consistency-aware update.} Robust 3D instances with enriched semantics preserve consistency under dynamic scene changes, enabling accurate target identification and scene updates in clutter.
    }
    \label{fig:5}
    \vspace{-7mm}
\end{figure}

With an instance-level scene representation enriched by semantic information, the robot can follow language-grounded instructions to execute multi-stage grasping. \figref{fig:5} illustrates multi-stage target identification, showing how our framework supports the subsequent grasping actions.

\noindent
1) \textbf{Language-grounded Target Identification and Grasping. }
From the mask grouping stage, each group yields a set of per-view instance images. Each group is embedded with a multi-view image encoder of Duoduo CLIP, aggregating cross-view evidence to obtain more robust instance embeddings, particularly advantageous in cluttered scenes. Input images were preprocessed by replacing the background outside each mask with white. Given a text query, we compute its similarity to all instance embeddings to identify the target object. A 6-DoF grasp pose, enabling target-oriented grasping, is estimated using a grasp pose detector~\cite{wu2025economic}. Beyond target identification, these stored instance embeddings also enable instant subsequent procedure after scene updates.

\noindent
2) \textbf{Post-interaction Segmentation Consistency. }
Following interaction, we capture a single RGB image to update the scene state for multi-stage grasping. 
Grounded SAM produces new noisy mask candidates, which are matched with existing instances via hierarchy-based instance grouping (\secref{sec:inst-seg}), leveraging stored instance-level embeddings to preserve segmentation consistency. This segmentation consistency enables the subsequent consistency-aware update and maintains reliable correspondences for succeeding grasps.

\noindent
3) \textbf{Change Detection and Instance-wise Update. }
Since instance correspondences are preserved between pre- and post-interaction, we detect displaced instances from a single post-interaction RGB image. For each instance, we project its pre-interaction point cloud onto the post-interaction image plane and compute IoU with the new mask. Instances with $\mathrm{IoU}<\tau_{\text{IoU}}$ are flagged as displaced. We then estimate a rigid-body transform $(\textbf{R},\mathbf{t})\in \text{SE}(3)$ that aligns the pre-interaction point cloud with the new observation:

\begin{equation}
\label{eq:optim}
\textbf{R}^*, \mathbf{t}^* = \underset{\textbf{R}\in \text{SO}(3),\, \mathbf{t}\in\mathbb{R}^3}{\arg\min} \mathcal{L}_{\text{total}}(\textbf{R}, \mathbf{t})
\end{equation}

The objective combines a chamfer, photometric, and task prior regularization loss.
\begin{equation}
\label{eq:loss}
\mathcal{L}_{\text{total}} = \lambda_{\text{chamfer}} \mathcal{L}_{\text{chamfer}} + \lambda_{\text{photo}} \mathcal{L}_{\text{photo}} + \lambda_{\text{reg,z}}\mathcal{L}_{\text{reg,z}}
\end{equation}

\emph{Chamfer Loss} is the bidirectional distance between the 2D contour of reprojected original 3D instance and the contour extracted from the post-update consistent masks. Let $P$ be the set of valid 2D projections onto the image plane and $C$ the set of boundary pixels from the binary instance mask. The loss is
\begin{equation}
\label{eq:chamfer}
\mathcal{L}_{\mathrm{chamfer}}
= \frac{1}{|P|}\sum_{\mathbf{p}\in P}\min_{\mathbf{c}\in C}\|\mathbf{p}-\mathbf{c}\|_2
+ \frac{1}{|C|}\sum_{\mathbf{c}\in C}\min_{\mathbf{p}\in P}\|\mathbf{c}-\mathbf{p}\|_2
\end{equation}
This provides stable gradients even under severe drift without intersection and enforces shape alignment.

The \emph{Photometric Loss} is formulated as a weighted L1 distance between each point's color and the image intensity at its projected location, enforcing appearance consistency once the silhouette is roughly aligned.
The \emph{Regularization Loss} penalizes excessive vertical translation, constraining \(z\)-motion while allowing planar movement.

For robustness, we adopt a coarse-to-fine schedule. In the coarse stage, we place a large weight on $\mathcal{L}_{\mathrm{chamfer}}$ to drive stable shape alignment. If the resulting IoU exceeds a threshold $\tau_{\text{IoU}}$, we terminate early, thereby skipping refinement and saving computation when alignment is already sufficient. Otherwise, we proceed to a fine stage that increases the weight for $\mathcal{L}_{\text{photo}}$ to refine with higher appearance fidelity.

Only displaced instances are selectively updated, while static ones are left unchanged. This observation-driven selective update keeps the scene representation synchronized with the latest observation across successive interactions, avoids redundant optimization and re-association for unchanged objects, and thereby enables fast multi-stage grasping.

\input{tab/real-inst-seg}

%% file: alg/algorithm1.tex
\newcommand{\LC}[1]{\Statex {\color{blue}\texttt{//}~#1}}
\newcommand{\LCi}[1]{%
  \Statex\hspace*{\algorithmicindent}{\color{blue}\texttt{//}~#1}%
}
\NewDocumentCommand{\LCindent}{O{1}m}{%
  \Statex 
  \hspace*{\dimexpr #1\algorithmicindent\relax}%
  {\color{blue}\texttt{//}~#2}\strut%
}

\newcommand{\LCii}[1]{\LCindent[2]{#1}}
\algrenewcommand\algorithmicrequire{\textbf{Input:}}
\algrenewcommand\algorithmicensure{\textbf{Output:}}


\begin{figure}[t]
\vspace{-6mm}
\begin{algorithm}[H]
\caption{Hierarchy-based Instance Mask Grouping}
\label{algo:1}
\begin{algorithmic}[1]
\footnotesize
\Require Per-frame Trees $\mathcal{T}=\{T_f\}$, Duoduo CLIP $\{\mathbf{c}_v\}$ 
\Ensure Grouped Leaf Graph $G$

\State $G \gets \textsc{ConstructLeafGraph}(\mathcal{T})$
\State $G \gets \textsc{GroupBySimilarity}(G,\mathbf{c})$
\State \Return $G$

\LC{Build leaf graph from frame-trees}
\Procedure{ConstructLeafGraph}{$\mathcal{T}$}
  \State $V \gets \bigcup_{f} \textsc{Leaves}(T_f)$;\; $G \gets (V,\emptyset)$
  \ForAll{$(u,v)\in \binom{V}{2}$ \textbf{with} $\phi(u)\neq \phi(v)$}
    \State $E_{u,v}\gets\{S_{\text{spatial}}(u,v),S_{\text{semantic}}(u,v)\}$
  \EndFor
  \State \Return $G$
\EndProcedure

\LC{Two-stage similarity grouping}
\Procedure{GroupBySimilarity}{$G,\mathbf{c}$}
  \LCi{Stage 1 (spatial similarity based grouping)}
  \While{$\exists (u,v)\in E(G): s_{\mathrm{spat}}(u,v)\ge \tau_{\mathrm{spat}}$}
    \State $(u^*,v^*)\gets \arg\max s_{\mathrm{spat}}(u,v)$
    \State \Call{GroupAndRewire}{$u^*,v^*$}
  \EndWhile
  \LCi{Stage 2 (semantic similarity based grouping)}
  \While{$\exists (u,v)\in E(G): s_{\mathrm{sem}}(u,v)\ge \tau_{\mathrm{sem}}$}
    \State $(u^*,v^*)\gets \arg\max s_{\mathrm{sem}}(u,v)$
    \State \Call{GroupAndRewire}{$u^*,v^*$}
  \EndWhile
  \State \Return $G$
\EndProcedure

\LC{Group and rewire}
\Procedure{GroupAndRewire}{$u,v$}
\State $w \gets u \cup v$ \Comment{Node Grouping}
\State $\textsc{AddNode(G,$w$)}$
  \ForAll{$x \in \big(\mathcal{N}(u)\cup\mathcal{N}(v)\big)\setminus\{u,v\}$} \Comment{Rewire}
    \State $s_{\mathrm{spat}} \gets \operatorname{mean}_{\substack{a\in\textsc{Set}(w)\\ b\in\textsc{Set}(x)}} S_{\text{spatial}}(a,b)$
    \State $s_{\mathrm{sem}} \gets \operatorname{mean}_{a,b}\,S_{\text{semantic}}(a,b)$
    \State $E_{w,x} \gets \{s_{\mathrm{spat}},\,s_{\mathrm{sem}}\}$
  \EndFor
  \State $\textsc{RemoveNode}(G,u);\textsc{RemoveNode}(G,v)$ 
\EndProcedure
\end{algorithmic}
\normalsize

\end{algorithm}
\normalsize
\vspace{-8mm}
\end{figure}

%% file: tab/real-inst-seg.tex
\begin{table*}[!t]
\centering
\caption{\textbf{Class-agnostic 3D instance segmentation} on the GraspClutter6D and our synthetic dataset. We compare Clutt3R-Seg with GraphSeg and MaskClustering. $\dagger$ means that we tuned the hyperparameters of both MaskClustering and GraphSeg, enhancing performance compared to their original versions on the GraspClutter6D dataset. Best results are highlighted in \textbf{bold}. Time is reported in seconds.}
\vspace{-1.5mm}
\label{tab:1}
\resizebox{\textwidth}{!}{
\begin{tabular}{c|ccc|ccc|ccc|c||ccc|c}
\toprule
\multicolumn{1}{c|}{} &
  \multicolumn{3}{c|}{Easy} &
  \multicolumn{3}{c|}{Intermediate} &
  \multicolumn{3}{c|}{Difficult} &
  \multicolumn{1}{c||}{\multirow{2}{*}{Time$\downarrow$}} &
  \multicolumn{3}{c|}{Synthetic} &
  \multirow{2}{*}{Time$\downarrow$}
  \\
Method &
  $\mathrm{AP}_{25}\uparrow$ &
  $\mathrm{AP}_{50}\uparrow$ &
  $\mathrm{AP}\uparrow$ &
  $\mathrm{AP}_{25}\uparrow$ &
  $\mathrm{AP}_{50}\uparrow$ &
  $\mathrm{AP}\uparrow$ &
  $\mathrm{AP}_{25}\uparrow$ &
  $\mathrm{AP}_{50}\uparrow$ &
  $\mathrm{AP}\uparrow$ &
    &
  $\mathrm{AP}_{25}\uparrow$ &
  $\mathrm{AP}_{50}\uparrow$ &
  $\mathrm{AP}\uparrow$ &
   
\\ \midrule
GraphSeg$\dagger$~\cite{tang2025graphsegsegmented3drepresentations} & 36.83& 10.72& 1.87& 34.60& 5.64& 0.97& 20.18& 2.42& 0.39& 71.9& 61.32& 9.89& 3.28& 51.9\\
MaskClustering$\dagger$~\cite{yan2024maskclusteringviewconsensusbased} & 59.59 & 5.21& 0.70 & 54.12 & 5.68 & 0.83 & 27.95 & 0.95 & 0.16 & \textbf{45.7} & 33.77 & 0.08 & 0.01 & 40.7 \\
\textbf{Clutt3R-Seg}  & \textbf{83.32}& \textbf{28.25}& \textbf{4.58}& \textbf{78.04}& \textbf{21.99}& \textbf{3.71}& \textbf{61.66}& \textbf{12.15}& \textbf{1.97}& 57.2& \textbf{94.40} & \textbf{64.78} & \textbf{16.06} & \textbf{21.5} \\
\bottomrule
\end{tabular}
}
\vspace{-5mm}
\end{table*}

%% file: 4_experiment.tex
\section{experiment}
\label{sec:experiment}

\subsection{Experiment Setup}
\noindent
\textbf{Datasets.} We evaluate our approach on the GraspClutter6D~\cite{back2025graspclutter6dlargescalerealworlddataset} dataset, containing real-world multi-view RGB-D sequences of high resolution of 1920$\times$1080, as well as our custom synthetic dataset generated in NVIDIA Isaac Sim~\cite{nvidia_isaac_sim}, with low resolution of 640$\times$480. Both datasets feature substantial clutter, heavy occlusion, and frequent inter-object contact. Our synthetic dataset consists of objects from the HouseCat6D~\cite{jung2024housecat6d}, containing 10 household categories, with various objects, backgrounds, and lighting conditions.

To systematically assess performance, we partition 99 GraspClutter6D sequences by the dataset-provided average visibility into three equally sized subsets {\textit{easy, intermediate, hard}} with mean overlap ratios of 11.62\%, 19.83\%, and 31.71\%, respectively.
For the sparse-view setting on the GraspClutter6D, we use only 8 posed images per sequence, selecting frames at approximately uniform intervals. For the synthetic dataset, we also use 8 images per scene by sampling camera poses reachable by a robot-mounted virtual camera in Isaac Sim, ensuring diverse and similarly displaced viewpoints within feasible working distances. 
Additionally, we evaluate the full pipeline on real-world data collected with a Franka Research 3 robot equipped with an Intel RealSense D435i, demonstrating successful on-hardware execution in real grasping scenarios. Qualitative results are provided in the supplementary video.

\noindent
\textbf{Evaluation metrics.} 
We evaluate the average precision (AP) at IoU thresholds of 25\% and 50\% (AP@25, AP@50), as well as the mean AP averaged over thresholds from 50\% to 95\% in 5\% increments. 
We consider two evaluation setups: \textit{class-agnostic instance segmentation}, which measures the accuracy of predicted instance regions in 3D regardless of category, and \textit{semantic instance segmentation}, which additionally evaluates consistency with semantic labels. 
For the evaluation of language-grounded semantic segmentation, IoU is computed over five visible object categories in 3D, all over identical 10 mm voxel grids for fair comparison across all methods. 
We use prompts corresponding to the five object categories defined in the GraspClutter6D dataset and additionally report the total computation time aggregated over all queries. The computation time for all baselines includes every module from preprocessing to completion.

\noindent
\textbf{Implementation details.} 
Since the input point cloud is reconstructed from predicted depth, we apply noise-robust voxel aggregation at 5 mm with a minimum occupancy of $m{=}3$ per voxel. 
For super-voxel construction \secref{method:pre}, we use the 5 mm downsampled point cloud with parameters $\alpha{=}0.5, \beta{=}1.0$. Neighboring voxels with $d_{ij}\leq \tau_\text{merge}{=}0.01$ are then contracted into compact super-voxels.

For all experiments, we set $\tau_\text{spat}{=}0.5$, $\tau_\text{sem}{=}0.65$, and $\tau_\text{IoU}{=}0.75$. During the update stage, we optimize a two-stage objective. In stage 1, we use $\lambda_{\text{chamfer}}{=}50.0$, $\lambda_{\text{photo}}{=}0.5$, and $\lambda_{\text{reg,z}}{=}10.0$, while in stage 2 we adjust the weights to $\lambda_{\text{chamfer}}{=}10.0$, $\lambda_{\text{photo}}{=}2.0$, and $\lambda_{\text{reg,z}}{=}1.0$. All baselines and our method are evaluated on a single NVIDIA RTX 3090. 


\subsection{Class-agnostic 3D Instance Segmentation}
For class-agnostic 3D instance segmentation, we select three recent representative baselines: SAI3D~\cite{yin2024sai3d}, GraphSeg \cite{tang2025graphsegsegmented3drepresentations} and MaskClustering \cite{yan2024maskclusteringviewconsensusbased}. GraphSeg aligns with our sparse-view, training-free pipeline, while SAI3D and MaskClustering serve as widely adopted state-of-the-art references. For fair comparison, we standardize the depth backend by replacing GraphSeg’s original DUSt3R~\cite{wang2024dust3r} with MVSAnywhere~\cite{izquierdo2025mvsanywhere}, the same estimator used in our pipeline, slightly improving its accuracy.

\begin{figure*}[t]
    \centering
    \includegraphics[width=0.8\textwidth]{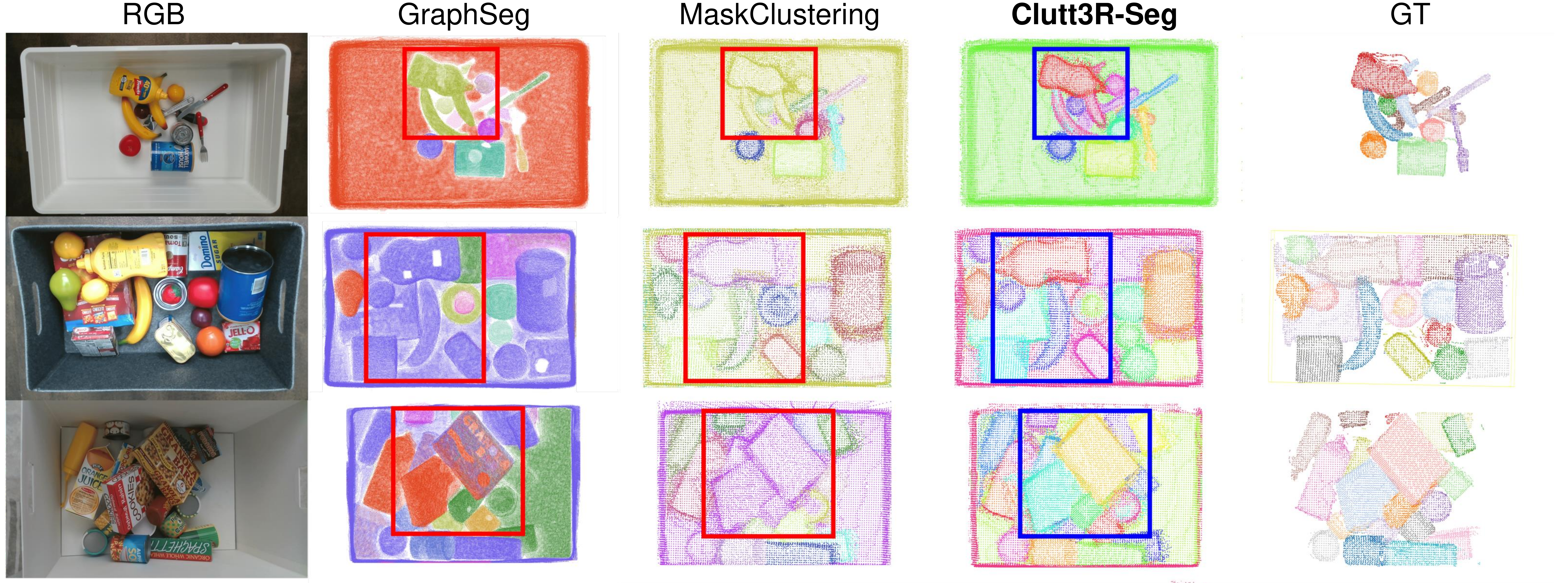}
    \caption{Qualitative result of GraspClutter6D dataset. The \textcolor{red}{red box} in baselines indicates erroneous segmentation, while the \textcolor{blue}{blue box} on Clutt3R-Seg shows correct segmentation. 1$^{\text{st}}$ row: easy sequence, 2$^{\text{nd}}$ row: intermediate sequence, 3$^{\text{rd}}$ row: difficult sequence. Clutt3R-Seg shows correct segmentation of different instances, whereas GraphSeg and MaskClustering both lead to severe under-segmentation. 
    }
    \label{fig:6}
    \vspace{-4mm}
\end{figure*}

\begin{figure*}[t]
    \centering
    \includegraphics[width=0.8\textwidth]{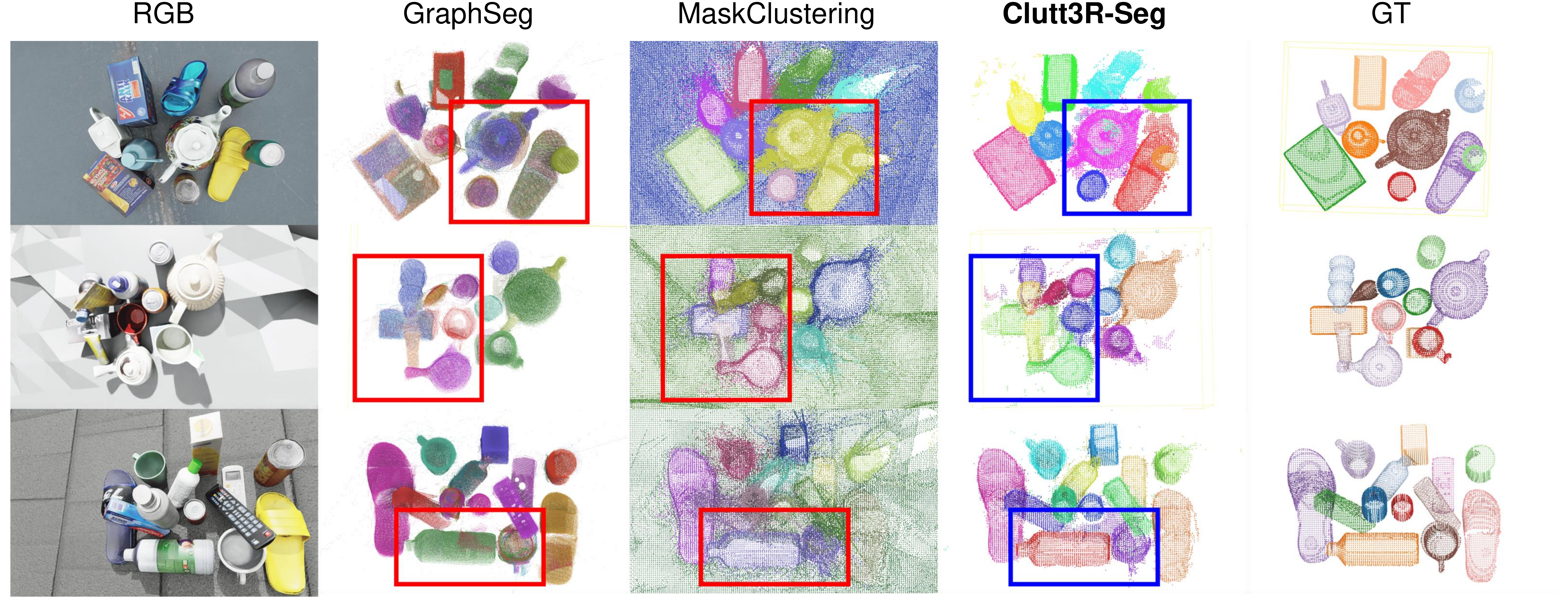}
    \caption{Qualitative result of our synthetic dataset. The \textcolor{red}{red box} in baselines indicates erroneous segmentation, while the \textcolor{blue}{blue box} on Clutt3R-Seg shows correct segmentation. The under-segmentation issue in baselines is consistent, with Clutt3R-Seg showing robust performance.}
    \label{fig:7}
    \vspace{-4mm}
\end{figure*}

As shown in \tabref{tab:1}, Clutt3R-Seg substantially outperforms both baselines in all metrics and difficulties. It maintains strong performance as difficulty increases, unlike the baselines whose scores drop sharply, showing up to 2.2 times higher AP@25 in difficult sequences compared to MaskClustering. Qualitatively, our method shows robust instance segmentation under significant clutter and occlusion for both real and synthetic, illustrated in \figref{fig:6} and \figref{fig:7}.

In the case of SAI3D, we exclude it from all related comparisons (\tabref{tab:1}, \figref{fig:6} and \figref{fig:7}) because its cross-view affinity requires co-visibility and discards invalid views, leading to an unreliable affinity matrix under sparse-view, cluttered scenes.
By contrast, the other two baselines leverage graph-based multi-view fusion and are therefore more competitive under our evaluation protocol. Specifically, MaskClustering weights edges by a view-consensus rate, yet under sparse-view occluded scenes, the small observer set makes containment-based support unreliable, biasing the global mask graph and reducing AP. 
GraphSeg frequently under-segments in occluded regions, since its 2D edge addition depends on a fixed threshold, while 3D structural edges fail to add under partial observations.
Conversely, on highly textured objects (\figref{fig:6}, 3$^{\text{rd}}$ row), GraphSeg over-segments because the upstream 2D segmentation yields small, noisy fragments and the chamfer distance in 3D often fails to reliably merge the over-segmented masks.

In contrast, our hierarchy-based framework is structurally robust to under-segmentation via bottom-up grouping, while over-segmented fragments are corrected by parent-node substitution, enabling effective and robust cross-view correspondences. Additionally, weighting confidence by cluster size reduces the impact of random mask errors, making the method more reliable in cluttered scenes with a few views.

\input{tab/semantic-seg}


\begin{figure*}[t!]
\centering\includegraphics[width=0.82\textwidth]{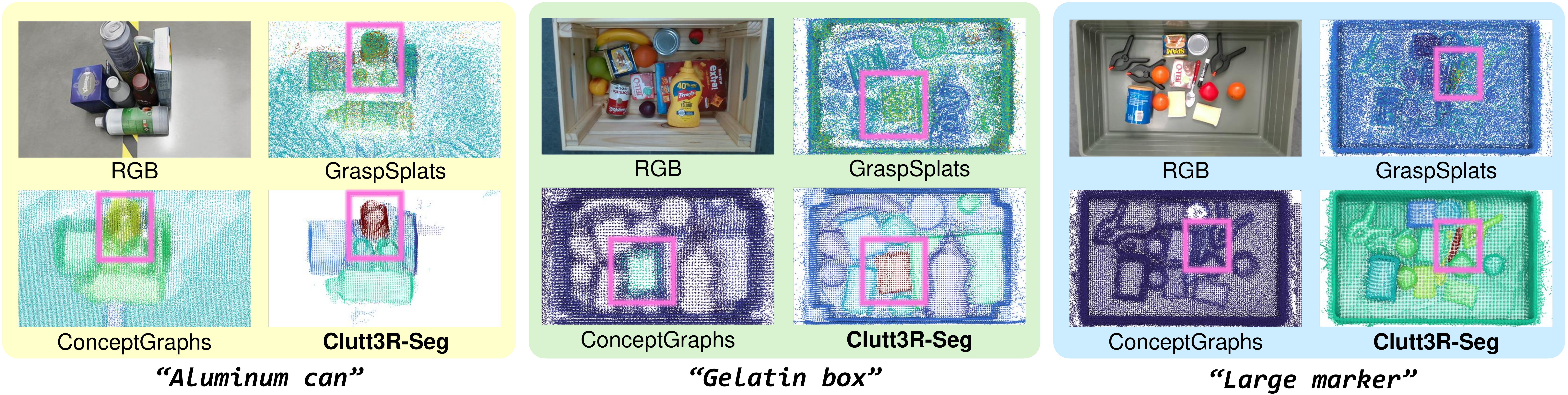}
    \caption{Qualitative result of language-grounded semantic segmentation. \textbf{Left}: synthetic dataset. \textbf{Middle}, \textbf{Right}: GraspClutter6D dataset. \textcolor{red}{Red} color indicates high similarity, and target object is highlighted in the \textcolor{magenta}{pink box}.}
    \label{fig:8}
    \vspace{-7mm}
\end{figure*}

\subsection{Language-grounded 3D Semantic Segmentation}

We compare Clutt3R-Seg with two representative baselines for language-grounded 3D semantic segmentation; GraspSplats~\cite{ji2024graspsplatsefficientmanipulation3d}, employing Gaussian splatting with language features, and ConceptGraphs~\cite{gu2024conceptgraphs}, constructing open-vocabulary 3D scene graphs from multi-view segmentation.
For fairness, we also include Clutt3R-Seg$^{-}$, where Duoduo CLIP is replaced with the original CLIP~\cite{radford2021learning}, aligning with the backbone used in the baselines.

As shown in \tabref{tab:2}, Clutt3R-Seg achieves the best performance with a 3D IoU of 52.54\%, demonstrating more than 50\% improvement over the baselines.
Notably, even Clutt3R-Seg$^{-}$ gains about 20\% over the baselines by simply averaging CLIP embeddings from multiple views.
Clutt3R-Seg further achieves the shortest runtime among all methods, demonstrating its efficiency. The improvements are due to the enrichment of semantic instances via enforcing cross-view consistency through hierarchy-based instance mask grouping, allowing Duoduo CLIP’s viewpoint-robust embeddings to support reliable language grounding even under severe clutter and occlusion as shown in \figref{fig:8}.

\subsection{Ablation Studies}
We analyze the role of the spatial similarity and residual node substitution on the GraspClutter6D dataset, as shown in \tabref{tab:3}.
Removing the spatial similarity reduces accuracy, highlighting the effectiveness of our two-stage grouping and its ability to resolve ambiguities that cannot be addressed by semantic features alone. 
Without substitution, residual nodes from over-segmentation remain unresolved, leading to sharper AP drops as the threshold increases from AP@25 to AP@50 and AP. This effect is amplified under sparse-view inputs, where missing observations result in greater geometric information loss, making substitution critical for maintaining accuracy at stricter evaluation levels. Substitution adds less than 0.5 s of overhead, making it a lightweight solution. Time in both tables is reported in seconds.

We further analyze viewpoint sparsity in \tabref{tab:4}. As the number of input views decreases, the overall AP declines, yet our method consistently outperforms all baselines across every metric, even when restricted to only four views. 
Additionally, runtime decreases with fewer views, suggesting that the system can be configured in a more lightweight mode at the expense of some performance.

\input{tab/ablation}
\input{tab/ablantion_view}

%% file: tab/semantic-seg.tex

\begin{table}[t]
\centering
\caption{\textbf{Quantitative evaluation of language-grounded 3D semantic segmentation} on the GraspClutter6D dataset. We report the 3D IoU scores (\%). Best results are highlighted in \textbf{bold}. Time is reported in seconds, and $^{-}$ indicates results obtained by replacing Duoduo CLIP with CLIP for a fair comparison.}
\label{tab:2}
\resizebox{0.90\columnwidth}{!}{%
\begin{tabular}{l|cc|cc}
\toprule
& \multicolumn{2}{c|}{GraspClutter6D} & \multicolumn{2}{c}{Synthetic} \\
Method & IoU$\uparrow$ & Time$\downarrow$ & IoU$\uparrow$ & Time$\downarrow$ \\
\midrule
GraspSplats~\cite{ji2024graspsplatsefficientmanipulation3d} & 17.38 & 81.1   & 20.70& 67.5\\
ConceptGraphs~\cite{gu2024conceptgraphs} & 32.59 & 148.5  & 34.90& 91.2\\
\textbf{Clutt3R-Seg}$^{-}$ & 50.61 & 47.6   & 41.69 & \textbf{23.7} \\
\textbf{Clutt3R-Seg} & \textbf{52.54} & \textbf{47.2} & \textbf{55.10}& 24.8\\
\bottomrule
\end{tabular}%
}
 \vspace{-7mm}
\end{table}

%% file: tab/ablation.tex
\begin{table}[t]
\centering
\caption{\textbf{Ablation study on the effectiveness} of our algorithm on the GraspClutter6D dataset. Best results are highlighted in \textbf{bold}.}
\label{tab:3}
\resizebox{0.8\columnwidth}{!}{%
\begin{tabular}{l|c|c|c|c}
\toprule
Method & $\mathrm{AP}_{25}\uparrow$ & $\mathrm{AP}_{50}\uparrow$ & $\mathrm{AP}\uparrow$ & Time $\downarrow$ \\
\midrule
w/o $S_{spatial}$ & 66.77& 16.33& 2.61& \textbf{55.2}\\
w/o substitution & 74.10& 20.60 & 3.19& 57.0\\
\textbf{Clutt3R-Seg}  & \textbf{74.34}& \textbf{20.79}& \textbf{3.41} & 57.2\\
\bottomrule
\end{tabular}%
}
\vspace{-3mm}
\end{table}

%% file: tab/ablantion_view.tex
\begin{table}[t]
\centering
\caption{\textbf{Ablation study on the number of viewpoints} on the GraspClutter6D dataset. Best results are highlighted in \textbf{bold}.}
\label{tab:4}
\resizebox{0.90\columnwidth}{!}{
\begin{tabular}{l|c|c|c|c}
\toprule
 Method & $\mathrm{AP}_{25}\uparrow$ & $\mathrm{AP}_{50}\uparrow$ & $\mathrm{AP}\uparrow$ & Time $\downarrow$ \\
\midrule
GraphSeg (8 Views) & 30.54& 6.20& 1.08 & 71.9\\
MaskClustering (8 Views) & 47.22& 3.95& 0.56 & 45.7\\
\textbf{Clutt3R-Seg} (4 Views) & 69.00& 16.54& 2.67 & \textbf{23.4}\\
\textbf{Clutt3R-Seg} (6 Views)  & 72.46& 19.46 & 3.15& 37.6\\
\textbf{Clutt3R-Seg} (8 Views)  & \textbf{74.34}& \textbf{20.79}& \textbf{3.41} & 57.2\\
\bottomrule
\end{tabular}
}
\vspace{-7mm}
\end{table}

%% file: 5_conclusion.tex
\section{Conclusion}
We introduce Clutt3R-Seg, a framework for sparse-view 3D instance segmentation for language-grounded multi-stage grasping in cluttered scenes. Leveraging a hierarchy-based instance mask grouping with substitution, our method achieves robust view-consistent segmentation and consistency-aware scene updates from a single post-interaction image. Experiments on both real and synthetic datasets, along with real-robot validation, demonstrate substantial improvements under heavy clutter reaching 61.66 AP@25, over 2.2× higher than the baselines.
Limitations remain due to predicted depth, which lowers reconstruction quality at stricter thresholds, and severe occlusion, which can prevent recovery of fully hidden objects or limit the completeness of single-view updates.
Future work will explore higher-fidelity depth estimation and multi-modal sensing to improve reconstruction and extend the framework to handle newly emerging objects in dynamic scenes.